\documentclass[10pt,twocolumn,letterpaper]{article}

\usepackage{cvpr}              

\usepackage{graphicx}
\usepackage{amsmath}
\usepackage{amssymb}
\usepackage{booktabs}
\graphicspath{ {./images/} }
\usepackage{color}
\usepackage[accsupp]{axessibility}  

\usepackage[pagebackref,breaklinks,colorlinks]{hyperref}

\usepackage[capitalize]{cleveref}
\crefname{section}{Sec.}{Secs.}
\Crefname{section}{Section}{Sections}
\Crefname{table}{Table}{Tables}
\crefname{table}{Tab.}{Tabs.}

\def\cvprPaperID{2915} 
\def\confName{CVPR}
\def\confYear{2023}

\newcommand{\MethodName}{AVFace\xspace}

\begin{document}

\title{AVFace: Towards Detailed Audio-Visual 4D Face Reconstruction}

\author{Aggelina Chatziagapi, Dimitris Samaras \\ \\
Stony Brook University $\;\;\;$\\
{\tt\small \{aggelina,samaras\}@cs.stonybrook.edu}}



\twocolumn[{%
\renewcommand\twocolumn[1][]{#1}%
\maketitle
\vspace{-20pt}
\centering
\includegraphics[width=\linewidth]{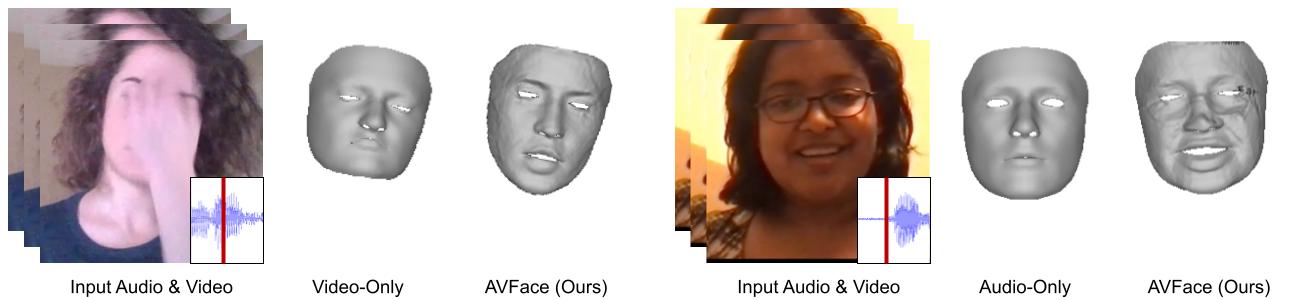}
\captionof{figure}{Given a monocular talking face video and the corresponding speech segment, \MethodName reconstructs the \textit{detailed 4D face geometry} for any input speaker, and it is robust in cases when either modality is insufficient.}
\label{fig:teaser}
\vspace{12pt}
}]

\begin{abstract}

In this work, we present a multimodal solution to the problem of 4D face reconstruction from monocular videos. 3D face reconstruction from 2D images is an under-constrained problem due to the ambiguity of depth. State-of-the-art methods try to solve this problem by leveraging visual information from a single image or video, whereas 3D mesh animation approaches rely more on audio. However, in most cases (e.g. AR/VR applications), videos include both visual and speech information. We propose \MethodName that incorporates both modalities and accurately reconstructs the 4D facial and lip motion of any speaker, without requiring any 3D ground truth for training. A coarse stage estimates the per-frame parameters of a 3D morphable model, followed by a lip refinement, and then a fine stage recovers facial geometric details. Due to the temporal audio and video information captured by transformer-based modules, our method is robust in cases when either modality is insufficient (e.g. face occlusions).
Extensive qualitative and quantitative evaluation demonstrates the superiority of our method over the current state-of-the-art.
Project page: \url{https://aggelinacha.github.io/AVFace/}.
\end{abstract}

\section{Introduction}
\label{sec:intro}


Reconstructing the 4D geometry of the human face has been a long standing research problem in computer vision and graphics. Accurate spatio-temporal (4D) face reconstruction has extensive applications in AR/VR, video games, virtual communication, the movie industry etc. However, recovering the per-frame 3D head pose and facial geometry from 2D images is an ill-posed problem due to the ambiguity of depth. Current approaches are largely based on 3D morphable models (3DMMs). Usually, they take a single image~\cite{DECA:Siggraph2021,EMOCA:CVPR:2022} or video~\cite{fml} as input and predict the 3DMM parameters. Some have also tried to predict additional geometric facial details, using the 3DMM fitting as prior~\cite{DECA:Siggraph2021,EMOCA:CVPR:2022,sider}. However, most of these video-only methods are either speaker-specific~\cite{sider,nha,gafni2021dynamic}, requiring to overfit to a specific speaker to recover their facial details, or fail to accurately capture fine details, like wrinkles and lip movements.
In addition, they cannot handle face occlusions, since they solely rely on the visual input. On the other hand, audio-driven 3D mesh animation approaches~\cite{VOCA2019,richard2021meshtalk,faceformer2022} learn better lip motion, but they require 4D ground truth scans for training, which are rare and expensive to capture. Furthermore, such audio-only methods cannot capture any speaker-specific characteristics or facial expressions, as they do not use any visual information. There is limited work in audio-visual 4D face reconstruction~\cite{moddropout,jointaudiovideo,realtime}, but these methods also require 4D ground truth scans and do not recover any facial geometric details.

In this work, we propose \MethodName that learns to reconstruct detailed 4D face geometry from monocular talking face videos, leveraging both audio and video modalities. Without requiring any 3D ground truth scans, it can recover accurate 4D facial and lip motion for any speaker. A coarse stage estimates a coarse geometry per frame, based on a 3DMM and using both image and speech features. Then, a SIREN MLP~\cite{siren} further improves the lip position, by learning an implicit representation of the lip shape conditioned on speech. Finally, a fine stage recovers geometric facial details, guided by pseudo-ground truth face normals and producing a high-fidelity reconstruction of the input speaker's face per frame. Due to the temporal audio and video information captured by transformer-based modules, 
our method is robust in cases when either modality is insufficient (e.g. face occlusions). To better handle such hard cases that are frequent in talking face videos, we further fine-tune our coarse stage with synthetic face occlusions.

In brief, the contributions of our work are as follows:
\begin{itemize}
    \item We propose \MethodName, a novel audio-visual method for detailed 4D face reconstruction, that follows a coarse-to-fine optimization approach, trained only on monocular talking face videos without any 3D ground truth.
    \item We introduce an audio-driven lip refinement network, and a fine stage guided by pseudo-ground truth face normals to accurately recover fine geometric details.
    \item Our temporal modeling, along with fine-tuning on synthetic face occlusions, makes our network robust to cases when either modality is insufficient.
\end{itemize}
\section{Related Work}
\label{sec:relatedwork}

\textbf{3D Face Reconstruction.}
Reconstructing the 3D face geometry from a single image or video has received a lot of attention in the last decades. One of the first methods was the work of Blanz and Vetter~\cite{blanz1999morphable}, which learns a 3DMM from 3D face scans. Following this work, several optimization-based~\cite{bas3dmm2017,gerig2018morphable,ploumpis2020towards} and learning-based~\cite{sota_facereconstruction,DECA:Siggraph2021,EMOCA:CVPR:2022,koizumi2020look,chang2018expnet,guo2020towards,kim2018inversefacenet,tran2017regressing,Yang_2020_CVPR,deng2019accurate} methods regress the parameters of a 3DMM from a single image. However, the PCA-based representation for shape and expression of 3DMMs is limited. Recently, deep learning approaches use the 3DMM fitting as prior and learn additional corrective displacements or normal maps~\cite{chaudhuri2020personalized,mesoscopicgeometry,3DFaceRecTIP18,cnnreconst,DECA:Siggraph2021,EMOCA:CVPR:2022,selfsupervised_ayush,wildphotobasedreconstruction,Richardson}, in order to capture missing facial details. DECA~\cite{DECA:Siggraph2021} and EMOCA~\cite{EMOCA:CVPR:2022} first predict the parameters of FLAME~\cite{FLAME}, and then learn a UV-map of vertex displacements.
Tran et al.~\cite{tran2017extreme} borrow details from reference bump maps in order to complete occluded facial regions.
Several subject-specific methods leverage multi-view images
~\cite{incrementalfacetracking,mesoscopicgeometry,deep_appearance_model,RingNet:CVPR:2019,realtime:cnn:animation,Liu_2018_CVPR,tran2017regressing,shang2020self} or learn implicit representations using multi-layer perceptrons (MLPs)~\cite{nha,sider,zheng2022imavatar,gafni2021dynamic,park2021nerfies}. However, these methods cannot generalize to multiple identities. Monocular video-based optimization techniques~\cite{pablo_geomfromvideo,3Dfacerig16,fml,shi2014automatic,garrido2016reconstruction,garg2013dense} leverage the multi-frame consistency to learn facial details. FML~\cite{fml} proposes a self-supervised method to learn identity and appearance models from videos. 
However, all these video-only methods ignore the audio modality that is usually included in most cases. In addition, they do not explicitly handle occlusions that frequently appear in talking face videos.

\textbf{Audio-Driven 3D Face Animation.} 
Another line of work addresses the problem of audio-driven 3D facial animation. Busso et al.~\cite{busso2005natural,busso2007rigid} use hidden Markov models (HMMs) to model the temporal relation between prosodic audio features and head motion sequences. Several works learn to animate subject-specific face models~\cite{karras2017audio,pham2017speech,richard2021audio,cao2005expressive} or artist-designed character rigs~\cite{taylor2017deep,edwards2016jali,zhou2018visemenet} based on input speech. More recently, VOCA~\cite{VOCA2019} maps audio features to 3D vertex displacements from a neutral face mesh. MeshTalk~\cite{richard2021meshtalk} aims to synthesize accurate lip motion and at the same time, plausible animation of parts of the face that are uncorrelated to the audio signal. FaceFormer~\cite{faceformer2022} proposes a transformer-based model that encodes long-term audio context and autoregressively predicts a sequence of 3D face meshes. While these works learn well-synchronized lip movements with the input speech, they cannot capture speaker-specific characteristics or facial expressions, since they do not use any visual input. Furthermore, they require high-quality 4D scans for training. In contrast, our method is trained on monocular talking face videos, without using any 3D ground truth.

\textbf{Audio-Visual 4D Face Reconstruction.}
There is limited work for 4D face reconstruction that considers both audio and video modalities. In~\cite{realtime}, the authors propose an audio-visual speaker-independent system for real-time facial tracking and animation that is robust to occlusions. Their approach is highly data-driven, using a pre-captured database of 3D mouth shapes, and requires depth information. Chen et al.~\cite{jointaudiovideo} present an optimization-based approach that learns a phoneme to 3D blenshape mapping, leveraging artist-designed blendshape models. An audio-visual deep learning approach is proposed by Abdelaziz et al.~\cite{moddropout}, where modality dropout encourages the network to pay attention to the audio input.
In contrast to all these methods, \MethodName does not require any 3D ground truth. It learns to recover detailed 4D geometry of any speaker and is robust to occlusions, trained only on monocular videos. 

\begin{figure*}[t]
  \centering
   \includegraphics[width=0.49\linewidth]{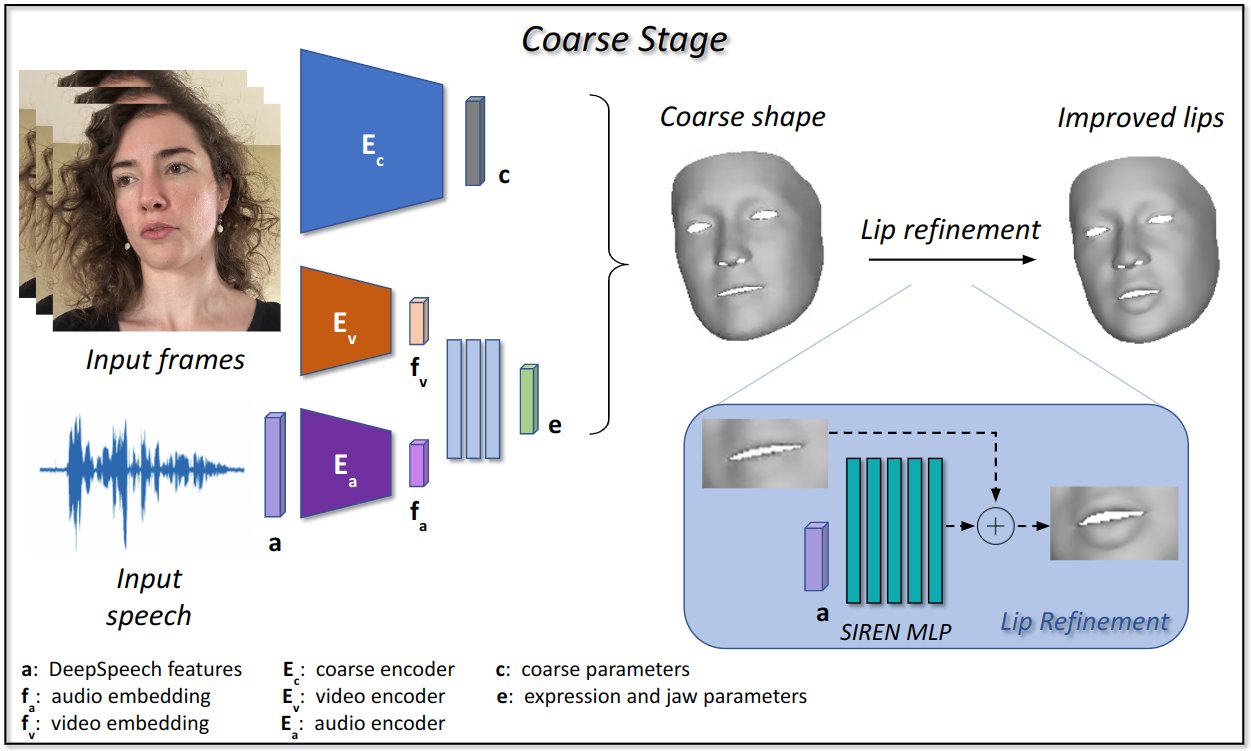}
   \includegraphics[width=0.49\linewidth]{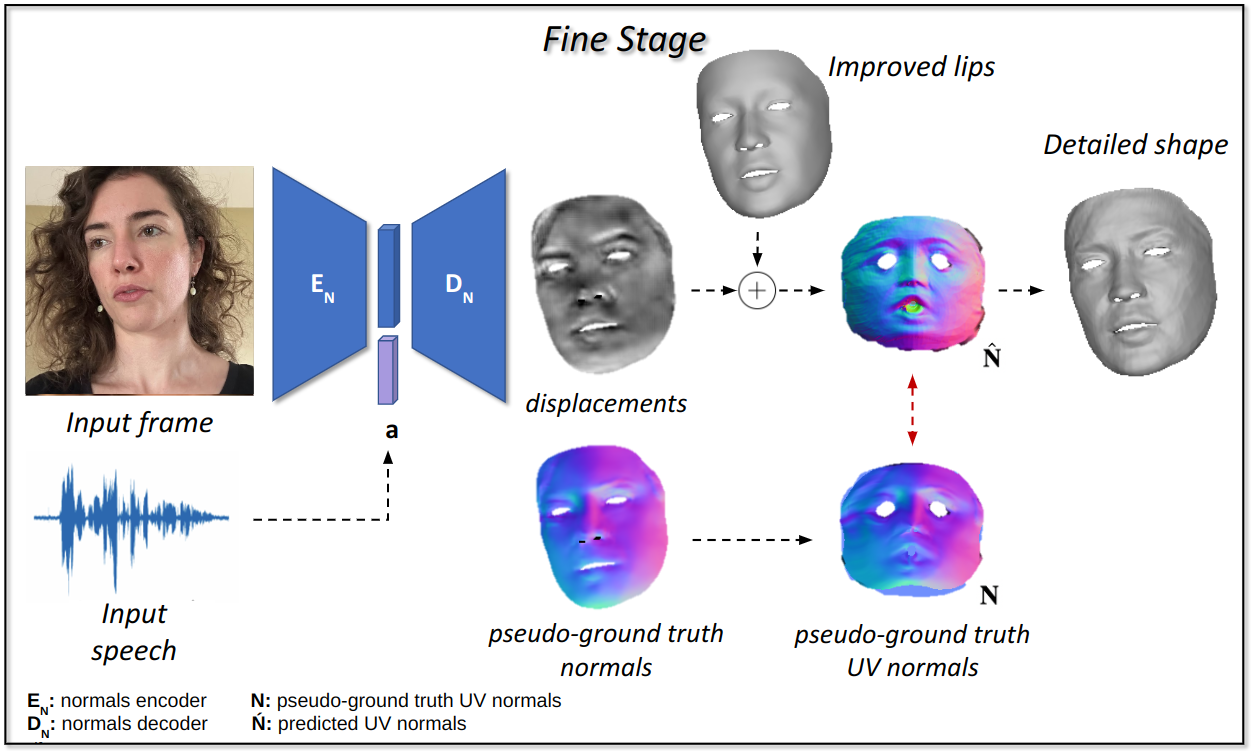}
   \caption{\textbf{\MethodName overview.} Given an input video of a talking face and the corresponding speech segment, \MethodName follows a coarse-to-fine optimization approach, in order to reconstruct the detailed 4D face geometry. A coarse stage estimates the per-frame parameters of a 3DMM. A SIREN MLP further improves the lip position. Finally, a fine stage recovers facial geometric details.}
   \label{fig:method}
\end{figure*}

\section{Method}

We present \MethodName, an audio-visual method that can accurately reconstruct the 4D face geometry of any input speaker. Given a monocular RGB video of a talking face and the corresponding speech segment, \MethodName follows a coarse-to-fine optimization approach, where audio and video modalities complement each other. Leveraging temporal information, it successfully handles occlusions (e.g. hand in front of the speaker's face) that are frequent in talking face videos. An overview of our method is shown in Fig.~\ref{fig:method}. In the following paragraphs, we describe each individual component in detail.

\subsection{Coarse Stage}

\textbf{Coarse Encoder.}
In the coarse stage, we first estimate a coarse geometry based on the FLAME morphable model~\cite{FLAME}. Given $x$ consecutive frames from an input video, a coarse encoder $E_{\text{c}}$ predicts a latent code $\mathbf{c} \in \mathbb{R}^{183}$ for each frame (see Fig.~\ref{fig:method}), which includes the FLAME parameters for head pose, camera, and shape, as well as the albedo and spherical harmonics coefficients for texture rendering~\cite{DECA:Siggraph2021}. Although we do not focus on texture prediction in this work, we use the rendered image for photometric loss during training. The coarse encoder consists of a ResNet-50~\cite{resnet} and a transformer encoder~\cite{attention}. The ResNet is initialized with pre-trained weights from DECA~\cite{DECA:Siggraph2021} and fine-tuned during training to predict intermediate embeddings of same size as $\mathbf{c}$. The transformer encoder is a stack of 3 encoder layers that include multi-head self-attention and feed-forward layers, to output the final $\mathbf{c}$ parameters per frame. Leveraging temporal information, it ensures smooth and accurate prediction of the head pose and camera for the sequence of $x$ input frames. In this way, we successfully handle cases where the face is partially or fully occluded, and single-image based methods like DECA~\cite{DECA:Siggraph2021} might fail.

\textbf{Expression and Jaw Prediction.} \MethodName includes a separate branch for the prediction of the jaw pose and expression parameters $\mathbf{e} \in \mathbb{R}^{53}$, based on both audio and video information. Unlike the expression encoder proposed in EMOCA~\cite{EMOCA:CVPR:2022}, our expression encoder has the following advantages: (a) we include both speech and visual information, handling cases where either modality is insufficient, (b) our transformer-based module captures temporal information across the $x$ input frames, compared to the single-image input of EMOCA, and (c) we include the jaw pose, which controls the mouth opening and closing and is crucial for accurate lip movement capture.

Given the raw audio signal of an input video, we extract DeepSpeech~\cite{deepspeech2} features $\mathbf{a} \in \mathbb{R}^{16 \times 29}$ per frame, which correspond to 29-dimensional embeddings for 16 overlapping windows of 20ms length each. DeepSpeech is a large RNN, trained on thousands of hours of data for automatic speech recognition. In this way, it captures useful information for any articulated phoneme and has been used for audio feature extraction by related works~\cite{thies2020neural,VOCA2019,guo2021ad}. The features $\mathbf{a}$ are given as input to an audio encoder $E_{\text{a}}$, which is an 1D convolutional network and learns an audio embedding $\mathbf{f_{a}}$ for each frame. A video encoder $E_{\text{v}}$ outputs the corresponding video embedding $\mathbf{f_{v}}$, following a ResNet-50 architecture. The audio and video embeddings are concatenated and passed to a transformer encoder of 3 stacked layers. Leveraging temporal information from the $x$ input frames, the transformer enhances the expression and jaw prediction. In cases of occlusion, the audio input leads to an accurate mouth position based on the recognized phoneme. In order to encourage the network to pay attention to the speech input, and not only to the visual input that is strongly correlated with the output shape, we use modality dropout during training, where $\mathbf{f_{a}}$ and $\mathbf{f_{v}}$ are zeroed out with a given probability~\cite{moddropout}.

\textbf{Training.} In the coarse stage, we minimize a similar objective with DECA, dubbed ${\mathcal{L}} _{\text{coarse}}$ loss~\cite{DECA:Siggraph2021}. In addition to the eye closure loss, we add a mouth closure and lip corner loss that measure the relative distances between upper/lower lips and left/right lip corners correspondingly~\cite{EMOCA:CVPR:2022}.



\subsection{Lip Refinement}\label{sec:liprefine}

\begin{figure}[t]
  \centering
   \includegraphics[width=\linewidth]{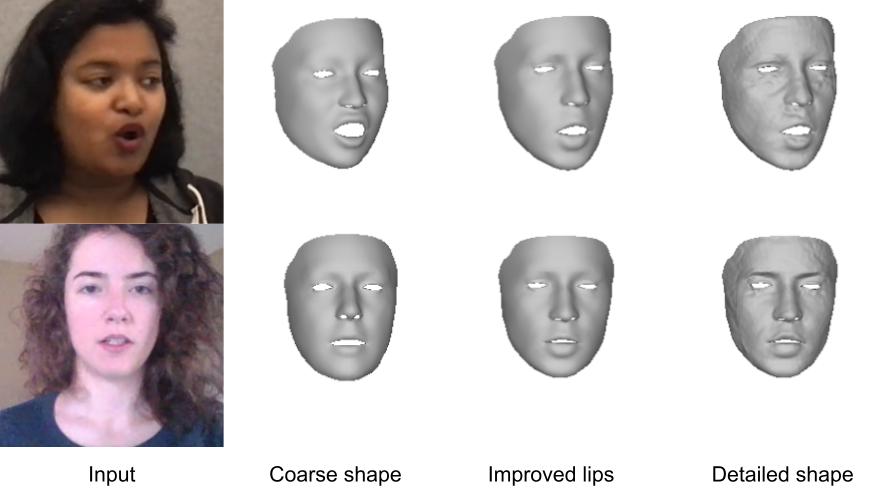}
   \caption{From the coarse stage of \MethodName, to the lip refinement, and finally to the detailed 3D reconstruction for an input frame.}
   \label{fig:stages}
\end{figure}

Following the coarse shape estimation, we further improve the lip shape during a lip refinement stage (see Fig.~\ref{fig:method}). For this purpose, we use a SIREN MLP~\cite{siren}, which can learn powerful implicit representations of complex natural signals~\cite{siren,chan2021pigan} and capture high frequencies due to the periodic activations. 
Conditioned on the audio features $\mathbf{a}$, the SIREN MLP predicts 3D vertex displacements for the coarse lip vertices, through a series of FiLM-conditioned linear layers~\cite{perez2018film}. We follow the architecture of the geometry network of NHA~\cite{nha}. However, in our case the FiLM layers are conditioned on the audio input and we predict lip offsets for any speaker, in contrast to the speaker-specific approach of NHA that predicts vertex offsets for the entire head conditioned on pose. The predicted lip offsets are added to the coarse lip vertices, in order to better capture the shape and position of the mouth. For example, in the first row of Fig.~\ref{fig:stages}, the lips of the coarse shape resemble an /a/ phoneme, and are modified to the correct /o/ phoneme during lip refinement (denoted as ``Improved lips"). The audio conditioning plays a significant role in this stage in inferring the lip offsets. For frames without audio, the lip offsets are negligible and the lip shape is deduced based on video only.

We train the lip refinement stage by minimizing the difference between the predicted normals in the lip region $\hat{N}_{\text{lip}}$ and the corresponding pseudo-ground truth normals ${N}_{\text{lip}}$ that are extracted from the pre-trained model for face normals by~\cite{Abrevaya_2020_CVPR}. To better capture the high frequencies, we apply the Laplace operator $\nabla^{2}$ as follows: 
\begin{equation}
    {\mathcal{L}} _{\text{lip}} = {\mathcal{L}} _{\text{coarse}} + \lambda_\text{n} \|\nabla^{2}\hat{N}_{\text{lip}} - \nabla^{2}N_{\text{lip}}\| _{1,1}
\end{equation}

\subsection{Fine Stage}

The fine stage recovers facial geometric details, like wrinkles and skin folds. In this stage, we train a convolutional neural network that predicts offsets of face normals from the coarse output as a one-channel image. The network is based on the encoder-decoder UNet-ResNet architecture proposed by~\cite{Abrevaya_2020_CVPR}, and it is initialized by their pre-trained weights. We experimentally found that adding audio input improves the normals prediction, especially in the mouth region. Thus, we concatenate the extracted audio features $\mathbf{a}$ as the last channel of the output of the encoder's third layer. The predicted normal displacements $D$ are combined with the coarse normals $N_c$ in the UV space:
\begin{equation}
    V_d = V_c + D \odot N_c
\end{equation}

\noindent
where $V_c$ and $V_d$ are the coarse and detailed vertices in the UV space correspondingly. The final detailed face normals $\hat{{N}}$, extracted from the detailed shape $V_d$, are learned by minimizing the normals loss, in order to capture high-frequency details, similarly to the lip refinement stage:
\begin{equation}
    {\mathcal{L}} _{\text{normals}} = \|\nabla^{2}\hat{N} - \nabla^{2}N\| _{1,1}
\end{equation}
\noindent
where ${N}$ are the pseudo-ground truth normals of the entire face, extracted from the pre-trained model by~\cite{Abrevaya_2020_CVPR}. We also minimize a photometric loss ${\mathcal{L}} _{\text{i}}$ on the final detailed rendered image, an ID-MRF loss ${\mathcal{L}} _{\text{mrf}}$~\cite{DECA:Siggraph2021, wang2018image}, a soft symmetry loss ${\mathcal{L}} _{\text{sym}}$~\cite{DECA:Siggraph2021}, and a regularization term ${\mathcal{L}} _{\text{reg}} = \|D\|_{1,1}$. Overall, the final loss of our fine stage is:
\begin{equation}
   {\mathcal{L}} _{\text{detail}} = {\mathcal{L}} _{\text{i}} + \lambda_\text{n} {\mathcal{L}} _{\text{normals}} + \lambda_{\text{m}}{\mathcal{L}} _{\text{mrf}} +\lambda_{\text{s}}{\mathcal{L}} _{\text{sym}} +  \lambda_\text{d}{\mathcal{L}} _{\text{reg}}
\end{equation}

\subsection{Fine-tuning with Synthetic Occlusions}\label{sec:occlusion}

Due to the temporal nature of our coarse stage and the audio input, \MethodName learns to handle occlusions, predicting the correct head pose, facial expression, and lip position in intermediate frames where part of the face is hidden, e.g. by a hand motion. Our training data include a small percentage of frames with face occlusions, while the speaker is talking. In order to enhance \MethodName's robustness in those cases, we fine-tune the coarse stage with a small number of synthetic data. We created frame sequences with synthetic hand occlusions, using the face occlusion generation method proposed by~\cite{voo2022delving}. For a random sequence of consecutive frames from our real data, we synthesize a hand occlusion for the first frame, transferring the color from the face via Sliced Optimal Transport~\cite{sot} (see examples in Fig.~\ref{fig:occlusions}). Then, we randomly rotate and translate the hand from frame to frame to approximate a realistic hand motion.

\subsection{Implementation Details}

We use the first 100 shape, 50 expression and 50 albedo parameters of FLAME, following DECA~\cite{DECA:Siggraph2021}. The transformer encoders for both the coarse and expression-jaw prediction include 3 encoder layers, 4 heads for the self-attention mechanism, 256 hidden units for the feed-forward layers, no dropout, and layer normalization~\cite{attention}, followed by 3 feed-forward layers with ReLU activations for projection.
During training, we use windows of $x=16$ frames. During test time, we use overlapping windows of the same size with a step of 1 frame. We experimentally chose a size of 64 for the audio and video embeddings, $\mathbf{f_{a}}$ and $\mathbf{f_{v}}$, correspondingly. The modality dropout probabilities are $0.4$ for audio and $0.5$ for video. The loss weights are set to $\lambda_\text{l} = 5\times 10^{-1}$, $\lambda_\text{c} = 10^{-4}$, $\lambda_\text{n} = 10$, $\lambda_\text{m} = 5\times 10^{-2}$, $\lambda_\text{s} = 5\times 10^{-3}$, $\lambda_\text{d} = 5\times 10^{-3}$. We use Adam optimizer~\cite{kingma2014adam} with a learning rate of $10^{-4}$ and weight decay of $5\times 10^{-4}$. See the suppl. material for more implementation details.
\section{Experiments}

\subsection{Datasets}

We use a publicly available dataset~\cite{voxceleb2} that includes videos from 6112 identities for training.
We use the proposed train-test split, and perform quantitative evaluation on the test set, which we call FaceSet. The videos are sampled at 25 fps and the audio at 16 kHz. For the synthetic data generation (see Sec.~\ref{sec:occlusion}), we randomly sampled 3k and 1k sequences of 16 frames from the train and test set respectively.

For testing, we use Multiface~\cite{wuu2022multiface}, in order to evaluate our 3D recontruction. Multiface includes videos of 13 identities in a multi-view capture stage, as well as the corresponding 3D ground truth meshes per frame. The videos are captured at 30 fps. For our case, only a single frontal camera view is passed as input to our model.

We additionally evaluate our method on 8 self-captured videos, captured in different environments and from different camera setups. We downsampled them to 25 fps, 640x480 video resolution and 16kHz audio sampling rate.

\begin{table}[t]
  \centering
  \tabcolsep=0.12cm
  \small
  \begin{tabular}{@{}l|ccccc@{}}
    \toprule
    \textbf{Method} & \textbf{NME $\downarrow$} & \textbf{AUC $\uparrow$} & \textbf{NME (lips) $\downarrow$} & \textbf{NME (occ) $\downarrow$} \\
    \midrule
    $\mathbf{e}$ from Video & 2.60 & 0.82 & 1.78 & 2.91 \\ 
    + Temporal & 1.98 & 0.82 & 1.33 & 2.68 \\ \hline
    $\mathbf{e}$ from Audio & 2.80 & 0.81 & 2.28 & 2.86\\ 
    + Temporal & 2.10 & 0.81 & 1.56 & 2.55\\ \hline
    $\mathbf{e}$ from AV & 2.91 & 0.81 & 2.05 & 3.02\\
    + Dropout & 2.14 & 0.82 & 1.65 & 2.79\\
    + Temporal & 1.73 & 0.83 & 0.99 & 2.42\\
    + Lip Refine & \textbf{1.63} & \textbf{0.84} & \textbf{0.91}& \textbf{2.39}\\
    \bottomrule
  \end{tabular}
  \caption{\textbf{Ablation study on the coarse stage.} Quantitative results for different variations of our coarse stage on the FaceSet (col. 1-3) and on the synthetic test set with occlusions (col. 4).}
  \label{tab:ablation}
\end{table}

\begin{table}[t]
  \centering
  \tabcolsep=0.12cm
  \small
  \begin{tabular}{@{}l|ccc@{}}
    \toprule
    \textbf{Audio Features} & \textbf{NME $\downarrow$} & \textbf{AUC $\uparrow$} & \textbf{NME (lips) $\downarrow$} \\
    \midrule
    Mel-spectrogram & 2.20 & 0.82 &	1.70 \\
    Wav2vec 2.0 & 1.66 & 0.83 & 1.11\\
    DeepSpeech (Ours) & \textbf{1.63} & \textbf{0.84} & \textbf{0.91} \\
    \bottomrule
  \end{tabular}
  \caption{\textbf{Ablation study on the audio features.} Quantitative results on the FaceSet for different audio features, namely mel-spectrogram~\cite{wav2lip}, wav2vec 2.0~\cite{baevski2020wav2vec}, and DeepSpeech~\cite{deepspeech2}.}
  \label{tab:audio}
    \vspace{-5pt}
\end{table}

\begin{table}[t]
  \centering
  \tabcolsep=0.12cm
  \small
  \begin{tabular}{@{}l|ccc@{}}
    \toprule
    \textbf{Method} & \textbf{NME $\downarrow$} & \textbf{AUC $\uparrow$} & \textbf{NME (lips) $\downarrow$} \\
    \midrule
    Without synthetic occlusions & 2.62 & 0.74 & 1.98 \\
    With synthetic occlusions & \textbf{2.39} & \textbf{0.76} & \textbf{1.73} \\
    \bottomrule
  \end{tabular}
  \caption{\textbf{Importance of synthetic occlusions.} Quantitative results on the synthetic test set with occlusions, without and with fine-tuning on synthetic face occlusions.}
  \label{tab:occ}
\end{table}


\begin{figure*}[t]
  \centering
   \includegraphics[width=\linewidth]{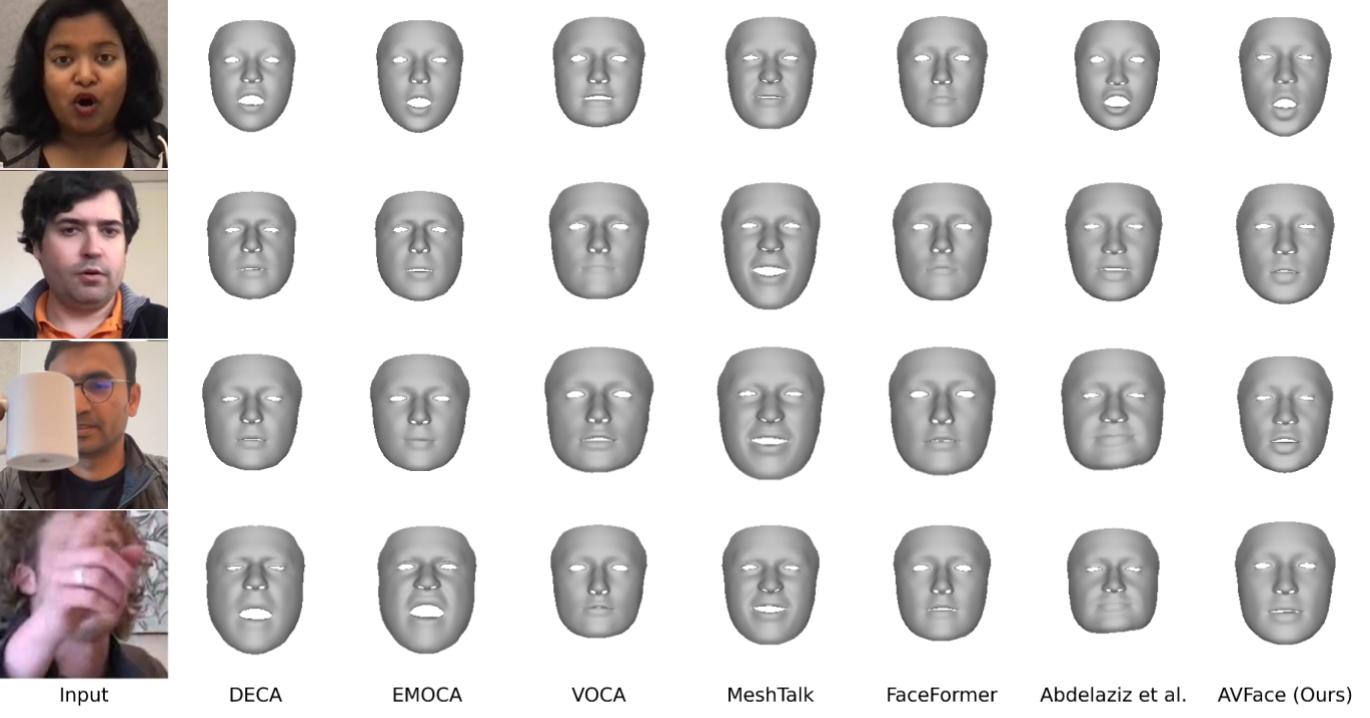}
   \caption{Comparison of our \textbf{coarse} reconstruction in frontal head pose with \textit{video-only} (DECA~\cite{DECA:Siggraph2021}, EMOCA~\cite{EMOCA:CVPR:2022}), \textit{audio-only} (VOCA~\cite{VOCA2019}, MeshTalk~\cite{richard2021meshtalk}, FaceFormer~\cite{faceformer2022}) and \textit{audio-video} (Abdelaziz et al.~\cite{moddropout}) methods.}
   \label{fig:audio}
\end{figure*}

\begin{figure*}[t]
  \centering
   \includegraphics[width=\linewidth]{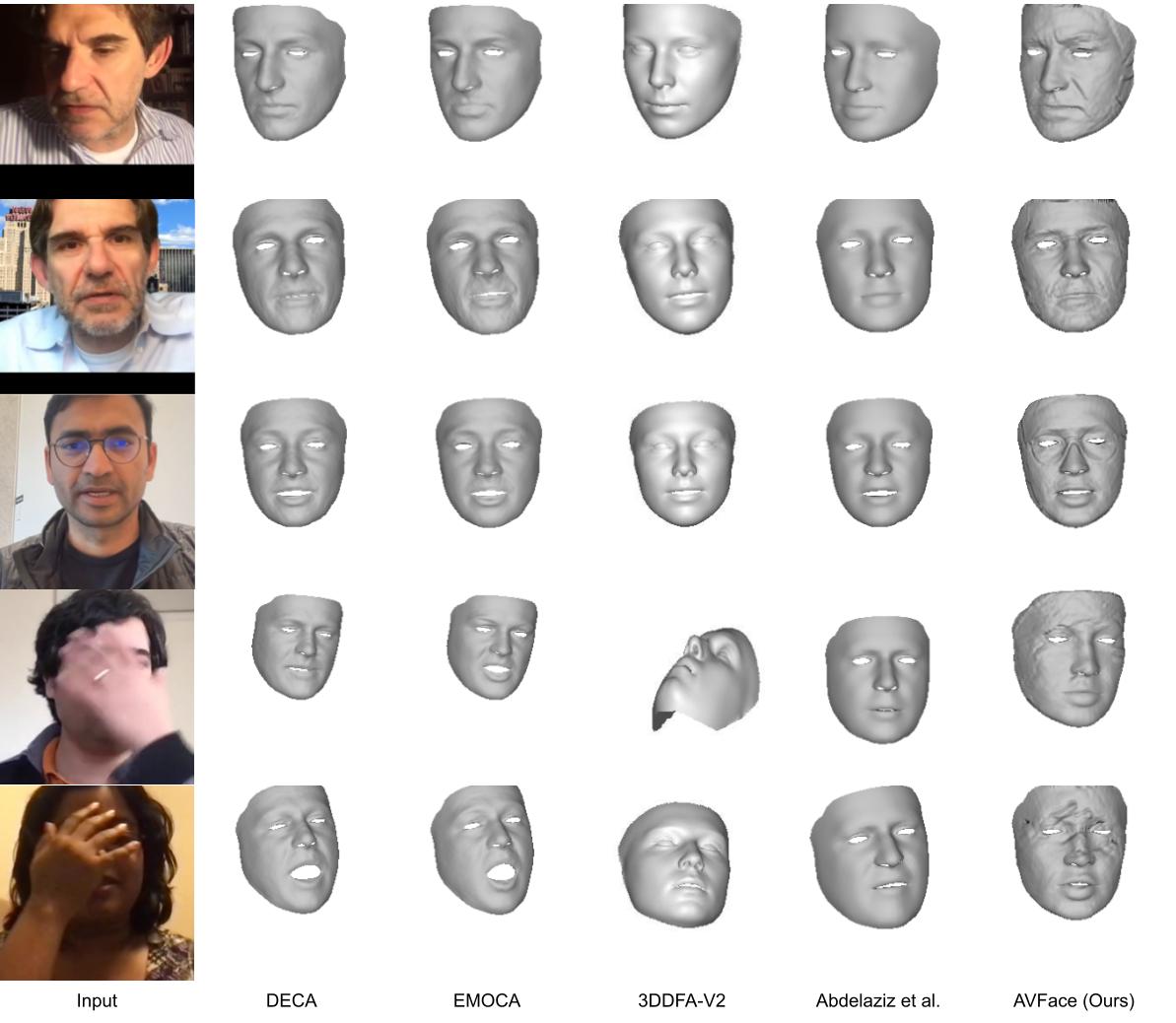}
   \caption{Comparison of our final \textbf{detailed} reconstruction with \textit{video-only} (DECA~\cite{DECA:Siggraph2021}, EMOCA~\cite{EMOCA:CVPR:2022}, 3DDFA-V2~\cite{guo2020towards,3ddfa_cleardusk}) and \textit{audio-video} (Abdelaziz et al.~\cite{moddropout}) methods.}
   \label{fig:pose_detail}
   \vspace{-10pt}
\end{figure*}

\subsection{Ablation Study}

We conduct an ablation study, in order to investigate the contribution of each part of our model. Fig.~\ref{fig:stages} shows the improvement from the coarse to the fine stage. The coarse stage produces a smooth 3D face geometry, capturing a rough estimate of the mouth position. During lip refinement, the lips are corrected. Finally, the fine stage recovers 3D geometric details of the entire face. 

\textbf{Coarse Stage.} Table~\ref{tab:ablation} shows the quantitative results for different variations of network. Since FaceSet does not include 3D ground truth, we compute the following metrics: normalized mean error (NME) (\%), area under the curve (AUC), NME computed only for the lips, and NME on the synthetic test set with occlusions. In the first and third rows, the jaw pose and expression parameters $\mathbf{e}$ are predicted using video only (only $E_\text{v}$) and audio only (only $E_\text{a}$) correspondingly. Temporal modeling boosts their performance. The fifth row considers both modalities and the sixth row adds the modality dropout. We suspect that the simple concatenation of audio and video embeddings confuses the network. Adding the modality dropout encourages the network to pay attention to the individual modalities and extract useful information, thereby improving the performance. Then, the addition of temporal information via the transformer-based modules and lastly the addition of lip refinement further improve the results.

\textbf{Audio Features.} We compared our audio feature extraction with (a) mel-spectrograms and (b) wav2vec 2.0~\cite{baevski2020wav2vec}. For mel-spectrograms, we use the proposed 2D convolutional audio encoder of~\cite{wav2lip} to get the audio embedding $\mathbf{f_{a}}$. Wav2vec 2.0~\cite{baevski2020wav2vec} is a state-of-the-art transformer-based ASR model, which learns powerful speech representations in a self-supervised manner. 
We use the pre-trained wav2vec 2.0 weights~\footnote{https://huggingface.co/facebook/wav2vec2-base-960h} and the final output is mapped to a 64-dimensional $\mathbf{f_{a}}$. The quantitative comparison is shown on Table~\ref{tab:audio}. The DeepSpeech features led to the best performance.

\textbf{Synthetic Occlusions.} 
Table~\ref{tab:occ} demonstrates the performance boost in cases of occlusion with the additional fine-tuning on our synthetic data (see Sec. ~\ref{sec:occlusion}).


\subsection{Qualitative Evaluation}

Fig.~\ref{fig:audio} compares our coarse reconstruction with state-of-the-art methods. Since there is limited work for joint audio-video 4D face reconstruction, we include comparisons with recent video-only, audio-only and audio-video approaches. For a fair comparison with the audio-only methods, which animate an input neutral mesh, we turn all results in frontal head pose. As input to the audio-only methods, we use a FLAME mesh fitted to the first frame of the video, in order to capture the speaker identity. As shown, the video-only methods DECA~\cite{DECA:Siggraph2021} and EMOCA~\cite{EMOCA:CVPR:2022} fail in case of occlusion, since they only consider a single image as input. The audio-only methods, namely VOCA~\cite{VOCA2019}, MeshTalk~\cite{richard2021meshtalk}, and FaceFormer~\cite{faceformer2022}, are not affected by occlusions, since they only consider audio input. Abdelaziz et al.~\cite{moddropout} have not published their code, so we implemented it ourselves. Overall, our method captures the lip motion more accurately than the other methods (see lips in first two rows) and is robust to occlusions (last two rows). For the last row where the face is fully occluded, the speaker pronounces the word /it/ at that specific moment, and only the lips of FaceFormer and our method resemble this sound.



\begin{table}[t]
  \centering
  \tabcolsep=0.12cm
  \small
  \begin{tabular}{@{}l|cccc@{}}
    \toprule
    \textbf{Method} & \textbf{NME $\downarrow$} & \textbf{AUC $\uparrow$} & \textbf{NME (lips) $\downarrow$} & \textbf{NME (occ) $\downarrow$} \\
    \midrule
    DECA & 1.85 & 0.82 & 1.25 & 3.89\\
    EMOCA & 1.98 & 0.80 & 1.53 & 3.89\\
    3DDFA-V2 & 1.64 & 0.84 & 1.15 & 3.47\\
    \midrule

    VOCA & 3.15 & 0.66 & 2.66 & 4.95\\
    MeshTalk & 4.48 & 0.47 & 4.59 & 4.89\\
    FaceFormer & 3.14 & 0.65 & 2.61 & 4.82\\
    \midrule
    
    Abdelaziz et al. & 2.26 & 0.82 & 1.72 & 2.76\\
    \MethodName (Ours) & \textbf{1.63} & \textbf{0.84} & \textbf{0.91}& \textbf{2.39}\\
    \bottomrule
  \end{tabular}
  \caption{Quantitative comparison of our method with state-of-the-art methods on the FaceSet (col.~1-3) and on the synthetic test set with occlusions (col.~4). The first 3 rows correspond to \textit{video-only} (DECA~\cite{DECA:Siggraph2021}, EMOCA~\cite{EMOCA:CVPR:2022}, 3DDFA-V2~\cite{guo2020towards,3ddfa_cleardusk}), the next 3 to \textit{audio-only} (VOCA~\cite{VOCA2019}, MeshTalk~\cite{richard2021meshtalk}, FaceFormer~\cite{faceformer2022}) and the last rows to \textit{audio-video} (Abdelaziz et al.~\cite{moddropout}) methods.}
  \label{tab:results}
\end{table}

Fig.~\ref{fig:pose_detail} compares our final detailed reconstruction with video-only and audio-video methods. Similarly, the video-only methods DECA~\cite{DECA:Siggraph2021}, EMOCA~\cite{EMOCA:CVPR:2022} and 3DDFA-V2~\cite{guo2020towards,3ddfa_cleardusk} fail in case of occlusion, both in terms of the predicted head pose and the lip movement. The audio input in Abdelaziz et al.~\cite{moddropout} helps the mouth prediction, but this method cannot handle occlusions either, since it lacks temporal modeling. Our method includes temporal audio and video information, and is fine-tuned to a small number of synthetic occlusions, making it robust in those cases. Furthermore,
\MethodName recovers accurate facial geometric details, like wrinkles and skin folds, that are largely missed by the other methods.

\subsection{Quantitative Evaluation}

Table~\ref{tab:results} shows the quantitative evaluation.
For the audio-only methods, we use the head pose as predicted by DECA. We notice that the audio-only methods perform poorly in terms of these metrics, as they do not consider visual information, and as a result they miss facial expressions and large mouth openings. The video-only methods give a high error in case of occlusions. Our method outperforms all these approaches across all metrics, and with a large margin for the test occlusions.

\begin{figure}[t]
  \centering
   \includegraphics[width=\linewidth]{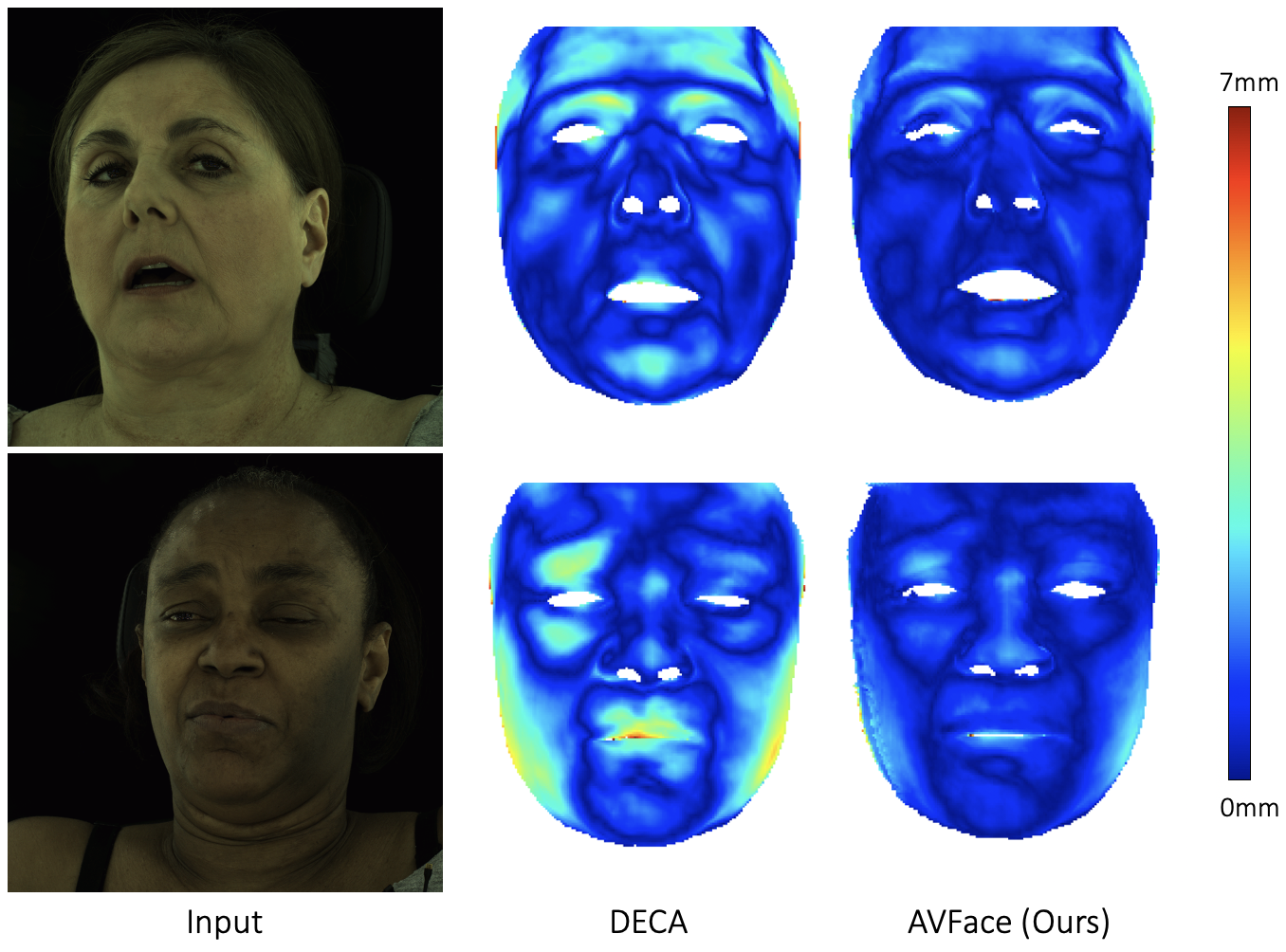}
   \caption{3D reconstruction error on Multiface data comparing with DECA~\cite{DECA:Siggraph2021}.}
   \label{fig:heatmap}
\end{figure}

To evaluate the entire 3D mesh, we additionally compute the 3D reconstruction error on Multiface data (see Table~\ref{tab:error}).
We follow the scan-to-mesh error computation of NoW~\cite{RingNet:CVPR:2019} and compare with the video-only and audio-video methods. Our method has the least error, and the error decreases by around $5\%$ from the coarse to the fine stage. Note that these data are not used for training, and they are very different from the training data in terms of illumination and camera pose. In addition, the ground truth meshes are relatively smooth, missing some fine details and as a result, the margin of our method from DECA is relatively small, in contrast to the qualitative comparison. Fig.~\ref{fig:heatmap} illustrates the 3D reconstruction error per vertex on the predicted mesh for 2 examples. We compare with DECA, since it gave the second lowest scores. Our method better reconstructs the speaker's face shape, facial details, as well as lip position.

\begin{table}[t]
  \centering
  \tabcolsep=0.12cm
  \small
  \begin{tabular}{@{}l|ccc@{}}
    \toprule
    \textbf{Method} & \textbf{Median (mm) $\downarrow$} & \textbf{Mean (mm) $\downarrow$} & \textbf{Std (mm) $\downarrow$} \\
    \midrule
    DECA & 1.68 & 2.01 & 1.60 \\
    EMOCA & 1.75 & 2.09 & 1.66\\
    3DDFA-V2 & 1.83 & 2.28 & 2.18\\
    Abdelaziz et al. & 1.71 & 2.06 & 1.67\\
    \MethodName (Ours) & \textbf{1.61} & \textbf{1.98} & \textbf{1.59}\\
    \bottomrule
  \end{tabular}
  \caption{3D reconstruction error on Multiface data comparing with \textit{video-only} (DECA~\cite{DECA:Siggraph2021}, EMOCA~\cite{EMOCA:CVPR:2022}, 3DDFA-V2~\cite{guo2020towards,3ddfa_cleardusk}) and \textit{audio-video} (Abdelaziz et al.~\cite{moddropout}) methods.}
  \label{tab:error}
  \vspace{5pt}
\end{table}


\begin{figure}[t]
  \centering
   \includegraphics[width=\linewidth]{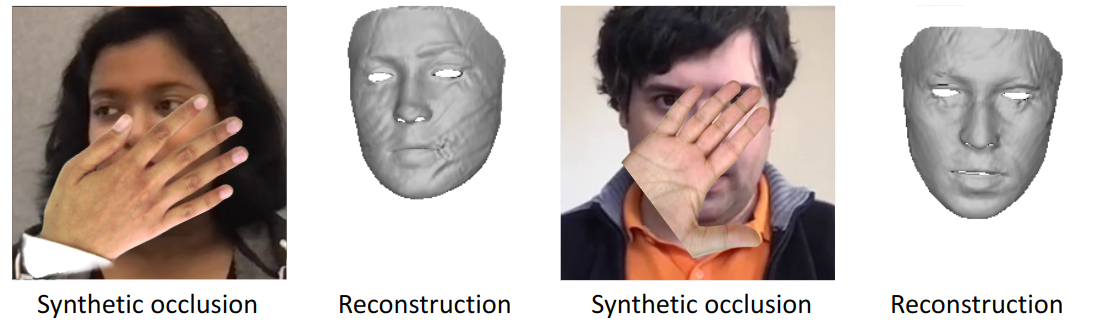}
   \caption{\textbf{Synthetic occlusions.} Example of our synthetic hand occlusions for a ground truth video frame and the corresponding 3D reconstruction.}
   \label{fig:occlusions}
\end{figure}

\begin{figure}[t]
  \centering
   \includegraphics[width=\linewidth]{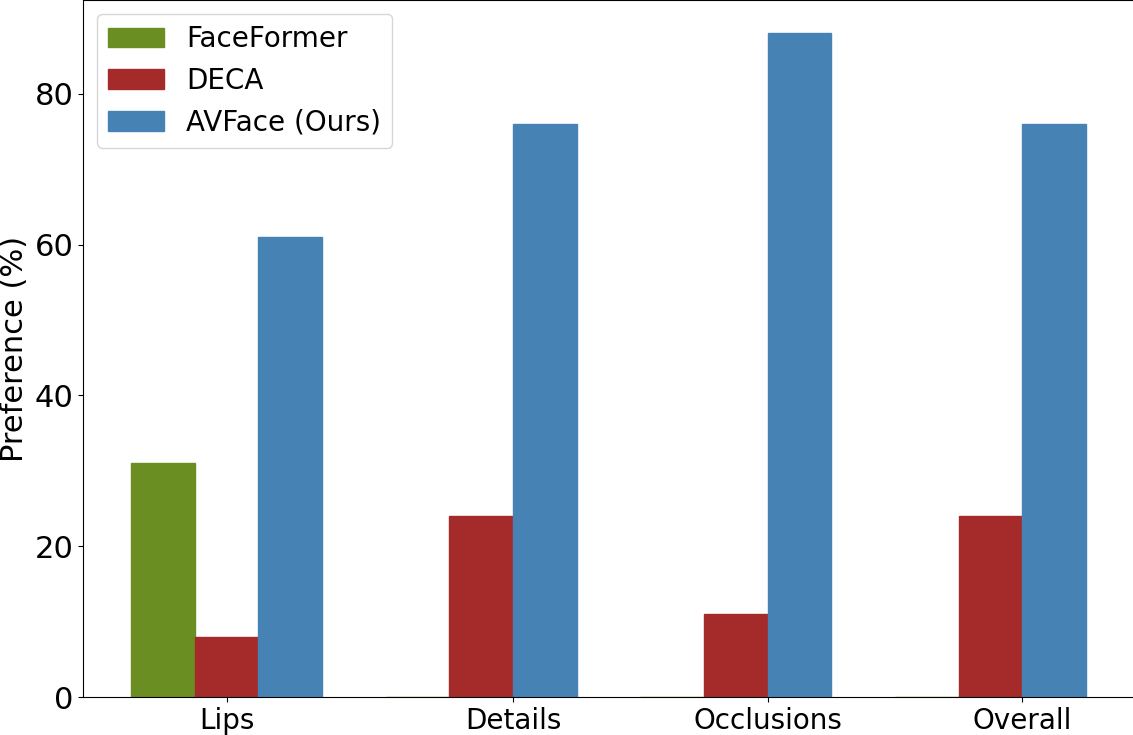}
   \caption{\textbf{User study.} Preference (\%) in terms of lip synchronization, facial details, robustness in occlusions, and overall quality, comparing FaceFormer~\cite{faceformer2022}, DECA~\cite{DECA:Siggraph2021}, and our method.}
   \label{fig:user}
   \vspace{-5pt}
\end{figure}

We additionally conducted a user study with 15 random participants to evaluate our 4D face reconstruction for full video sequences. We included 2 sections. The first section shows video results of FaceFormer~\cite{faceformer2022}, DECA~\cite{DECA:Siggraph2021}, and \MethodName in frontal head pose (similar to Fig.~\ref{fig:audio}) and ask the users to choose the method with most accurate lip synchronization (i.e. the lips better follow the speech). In the second section, we compare our final reconstruction with DECA, similar to Fig.~\ref{fig:pose_detail}. We ask the users to choose the method that (a) produces better facial details, (b) better handles face occlusions, and (c) has the best quality overall. For each section, we include 5 videos of around 10-30 seconds from FaceSet, making sure that they include face occlusions, are diverse w.r.t. speech variations (e.g. intonation, energy, speaking rate), and are balanced w.r.t. speaker age and gender. In this way, we expect to get a representative sample of talking face videos. The results of our user study are shown in Fig.~\ref{fig:user}. FaceFormer produces better audio-lip synchronization compared to DECA and a few users prefer DECA over \MethodName in terms of details and overall quality. In all questions, our method leads to the highest preference scores.

\section{Discussion}

\textbf{Limitations.}
Since we predict face normal offsets in the fine stage, face occlusions like hair, glasses and hands can be rendered as additional details, engraved on the speaker's face (see  Fig.~\ref{fig:occlusions}). This can be handled with additional temporal modeling or depth input which are out of the scope of this work. We also noticed that during occlusions of the whole face, our method might produce a more frontal head pose (last row in Fig.~\ref{fig:pose_detail}). In the future, we plan to further improve our method's robustness to such extreme cases.

\textbf{Ethical Considerations.}
We would like to note the potential misuse of face reconstruction methods.
With the advances in detailed 3D reconstruction and neural rendering, it becomes easier to generate photo-realistic fake videos of any speaker, which can be used for malicious purposes (e.g. to spread misinformation). Thus, it is important to ensure fair and safe use of videos, and develop accurate methods for fake content detection and forensics~\cite{yu2021artificial, wang2019cnngenerated}.

\section{Conclusion}

We propose \MethodName, a novel audio-visual approach that addresses the problem of 4D face reconstruction from monocular talking face videos.
Without requiring any 3D ground truth,
\MethodName recovers detailed 4D facial and lip motion
of any input speaker, following a coarse-to-fine optimization strategy. In addition, leveraging temporal information, it is robust to cases when either modality is insufficient (e.g. face occlusions).
Based on a thorough qualitative and quantitative evaluation, we show that \MethodName outperforms the current state-of-the-art. In the future, we plan to further enhance its robustness to extreme cases, where the face is fully occluded for multiple consecutive frames.


{\small
\bibliographystyle{ieee_fullname}
\bibliography{egbib}
}

\end{document}